%% file: root.tex
\documentclass[letterpaper, 10 pt, conference]{ieeeconf}  %

\IEEEoverridecommandlockouts                              %

\title{\LARGE \bf
Online Adaptation of Visual Odometry Frontends with Image-Conditioned Reinforcement Learning
}

\author{Simone Nascivera$^{1}$\orcidlink{0009-0008-2713-5534},
Leonard Bauersfeld$^{1}$\orcidlink{0000-0002-5790-9982},
Jeff Delaune$^{2}$\orcidlink{0000-0003-1509-4401}, and
Davide Scaramuzza$^{1}$\orcidlink{0000-0002-3831-6778}%
\thanks{$^{1}$Simone Nascivera, Leonard Bauersfeld, and Davide Scaramuzza are with
Robotics and Perception Group, University of Zurich, Switzerland.}%
\thanks{$^{2}$Jeff Delaune is with Jet Propulsion Laboratory,
California Institute of Technology, USA.}%
}

\usepackage{ral}
\usepackage{colortbl}

\hypersetup{
  breaklinks=true,
  colorlinks=true,
  citecolor=eccvblue,
  pdftitle={Online Adaptation of Visual Odometry Frontends with Image-Conditioned Reinforcement Learning},
  pdfauthor={Simone Nascivera, Leonard Bauersfeld, Jeff Delaune, Davide Scaramuzza},
  pdfsubject={Visual odometry and reinforcement learning},
  pdfkeywords={Visual odometry, Reinforcement learning, Adaptive parameter tuning, Feature tracking, Robotics}
}

\usepackage{orcidlink}

\usepackage[capitalise]{cleveref}

\begin{document}

\maketitle
\thispagestyle{empty}
\pagestyle{empty}

\input{sections/00_Abstract}

\input{sections/01_Introduction}
\input{sections/02_RelatedWork}
\input{sections/03_Methodology}

\input{sections/04_ExperimentalSetup}
\input{sections/05_Results}
\input{sections/06_Discussion}
\input{sections/A_Acknowledgements}

\input{root.bbl}
\end{document}

%% file: sections/00_Abstract.tex
\begin{abstract}
Visual odometry (VO) frontends are typically tuned offline by domain experts on
pre-recorded datasets and then deployed with fixed hyperparameters.  Yet a
configuration that performs best on a benchmark is not guaranteed to remain
best when texture, illumination, motion blur, sensor noise, or computational
conditions change at deployment.  We propose a frontend that instead adapts
its parameters automatically and continuously.  We formulate frontend tuning
as a sequential decision-making problem and introduce an image-conditioned
reinforcement-learning policy that combines a lightweight embedding of the
current image with a compact set of frontend statistics.  At each decision step, the policy
selects the FAST detection threshold, KLT patch size, and RANSAC rejection
threshold; a privileged critic provides additional context only during
training.  Trained on synthetic TartanAirV2 data, the policy transfers zero-shot
to a monocular-inertial OpenVINS pipeline on EuRoC, TUM-VI, and UZH-FPV. 
On synthetic test sequences, the learned policy improves the tracking-computation trade-off over an optimized static parameter configuration.  On the three real-world benchmarks, it reduces mean ATE by
up to $8\%$ and runtime by up to $57\%$ relative to static parameters baseline.  These results show that image-conditioned online
adaptation can improve the accuracy-computation trade-off beyond a single
configuration selected offline.
\end{abstract}

%% file: sections/01_Introduction.tex
\section{Introduction}
\label{sec:introduction}

Reliable localization in GPS-denied environments is a fundamental requirement
for autonomous robots operating indoors, underground, and in planetary
exploration settings~\cite{carlone2026slam,tranzatto2024team,zhao2024subt,
cheng2005visual,delaune2020xvio}.  Classical visual and visual-inertial odometry
(VO/VIO) is especially attractive on resource-constrained platforms.  Many
such systems detect sparse image patches and track them across frames using
Lucas--Kanade tracking (KLT)~\cite{mourikis2007msckf,forster2014svo,
bloesch2015robust,bloesch2017iterated,leutenegger2015keyframe,
geneva2020openvins,lucas1981klt}.

Despite their maturity, deploying these pipelines still relies heavily on
domain expertise.  Researchers and engineers tune feature-detection, tracking,
and outlier-rejection parameters offline on pre-recorded data, then freeze one
configuration for deployment.  A configuration that is best on a benchmark
such as TartanAir, EuRoC, or UZH-FPV~\cite{wang2020tartanair,
patel2025tartanground,burri2016euroc,delmerico2019uzhfpv}, however, is not
guaranteed to be best in the deployment environment.  Texture, illumination,
motion blur, sensor noise, and available computation may differ from the tuning
data and may change within a single trajectory.  Offline tuning therefore
selects a static compromise for a fundamentally time-varying problem.  This is
the dataset-to-reality gap we address.

\begin{figure}[t]
    \centering
    \includegraphics[width=\linewidth]{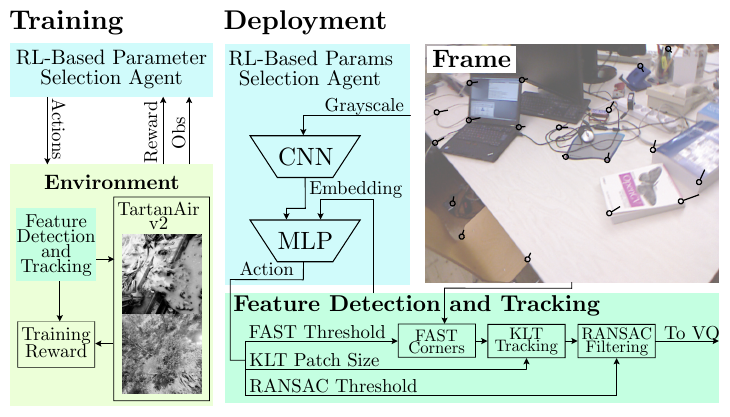}
    \caption{Our RL agent selects feature-detection, tracking, and outlier-
    rejection parameters online from an embedding of the current image and a
    compact frontend history.  Conditioning the decision on visual appearance
    allows the frontend to adapt to changes in texture density, motion blur,
    and sensor noise.}
    \vspace*{-12pt}
    \label{fig:figure1}
\end{figure}

Ideally, the frontend would instead tune itself automatically and continuously,
\emph{as if a human expert inspected every incoming image and selected the appropriate
configuration}.  We realize this idea by reframing frontend tuning as a
sequential decision-making problem and solving it with reinforcement learning
(RL).  Because images are too high-dimensional for a lightweight control policy,
we encode the current grayscale frame and combine its embedding with compact
frontend history.  From this state, the policy jointly selects the FAST
threshold, KLT patch size, and RANSAC threshold (Fig.~\ref{fig:figure1}).  The
policy is learned offline, but the frontend parameters continue to adapt online
as the image stream changes.

Existing adaptive frontends primarily use manually designed rules or internal
statistics~\cite{fontan2024anyfeature,yan2023plpf}.  RL-VO formulates VO
adaptation as sequential decision making~\cite{messikommer2024reinforcement},
but its policy does not observe the image and controls higher-level decisions
rather than detection and tracking parameters.  Our image-conditioned policy
can instead act on the visual causes of frontend degradation.  A lightweight
CNN provides the image embedding, while a privileged critic supplies additional
context only during training.

We train on TartanAirV2 and evaluate without target-domain tuning. On synthetic test sequences, the policy improves the tracking--computation trade-off over a static parameter configuration optimized with the same objective. At 20~Hz under the tighter computation budget, it reduces runtime by $16.3\%$, frames with fewer than 20 features by $73.8\%$, and deadline violations by $32.3\%$. When
integrated into monocular-inertial OpenVINS~\cite{geneva2020openvins}, it
transfers zero-shot to EuRoC, TUM-VI~\cite{schubert2018vidataset}, and UZH-FPV,
reducing mean ATE by $3.8$--$8.7\%$ and runtime by
$51.5$--$57.2\%$ relative to the static baseline. 
We chose OpenVINS because it uses the MSCKF-based estimator representative of the efficient filtering architectures deployed at scale on mobile devices; MSCKF-style VIO has been reported to underlie Google ARCore~\cite{geneva2020openvins,kim2022proprietary}.

In summary, our main contributions are:
\begin{enumerate}
    \item \textbf{Image-conditioned online tuning.} We formulate VO frontend
    configuration as a sequential decision-making problem in which a policy
    combines the current image embedding with compact frontend history and
    adapts feature detection, tracking, and outlier-rejection parameters online.
    \item \textbf{Efficient policy architecture.} We introduce a lightweight
    visual encoder and a privileged critic that provides additional context
    only during training, keeping the deployed policy suitable for
    resource-constrained platforms.
    \item \textbf{Synthetic-to-real validation.} We evaluate the policy under
    multiple frame rates and computation constraints and demonstrate zero-shot
    transfer from TartanAirV2 to three real-world VIO benchmarks, improving the
    accuracy--computation trade-off over a static baseline.
\end{enumerate}

%% file: sections/02_RelatedWork.tex
\section{Related Work}
\label{sec:related_work}

\subsection{Visual Odometry}
\label{ssec:visual_odometry}
Visual odometry (VO) pipelines are commonly categorized into classical and learned approaches based on their design. Classical methods, introduced more than three decades ago~\cite{cadena2016slam}, remain widely deployed in practice, ranging from virtual reality systems to drones~\cite{delaune2020xvio, zhou2015guidance} and smartphones. These pipelines are typically organized into tracking and mapping modules and are further grouped into direct, semi-direct, and indirect methods depending on how inter-frame correspondences are established.

\subsubsection{Classical Methods}
\label{sssec:classical_methods}
Direct methods estimate camera motion by optimizing photometric residuals directly on image intensities. Dense approaches use most available pixels~\cite{kerl2015dense}, whereas sparse direct methods such as DSO~\cite{engel2018directsparse} evaluate photometric residuals only around selected image points, reducing computational cost.

Sparse photometric alignment is also used as a correspondence mechanism in feature-based systems. For example, xVIO~\cite{delaune2020xvio} detects FAST keypoints and tracks their surrounding patches using KLT~\cite{lucas1981klt}. The resulting two-dimensional feature correspondences are subsequently incorporated as geometric measurements in an EKF. Thus, xVIO employs a sparse direct tracker, although its state estimator is feature based rather than a sparse direct photometric optimizer.

Feature-based methods, such as ORB-SLAM and its successors~\cite{murartal2015orb,murartal2017orbslam2,campos2021orbslam3}, estimate motion from geometric residuals on sparse image features, with correspondences commonly established through descriptor matching. Semi-direct approaches such as SVO~\cite{forster2014svo} combine sparse feature selection with photometric alignment of local patches. A related class of methods incorporates learned feature descriptors and matchers~\cite{detone2018superpoint,lindenberger2023lightglue,sarlin2020superglue} while retaining the classical feature-based pipeline structure.

\subsubsection{Learning-Based Methods}
\label{sssec:learning-based_methods}
In recent years, VO has increasingly adopted deep learning, reflecting broader trends in computer vision. End-to-end learned approaches directly regress camera motion from image streams~\cite{wang2017deepvo,wang2020tartanvo,wang2020tartanair}. While these methods can achieve strong performance within distributions similar to their training data, they often degrade under domain shift and exhibit limited out-of-distribution generalization.

Hybrid methods aim to combine the strengths of classical and learned approaches by preserving the overall pipeline structure while replacing individual components with neural networks. DROID-SLAM~\cite{teed2021droid} estimates dense correspondences and refines camera poses and depths through differentiable bundle adjustment. DPVO~\cite{teed2023dpvo} improves efficiency by replacing dense correspondence estimation with learned sparse patch tracking, while DPV-SLAM~\cite{lipson2024deeppatch} subsequently extends DPVO with loop-closure and global-optimization mechanisms. Although hybrid approaches improve robustness and interpretability compared to end-to-end methods, their learned components remain difficult to analyze and diagnose and are substantially more computationally demanding than lightweight methods such as SVO.

A recent trend explores the use of geometrically grounded foundation models within VO and SLAM systems. MASt3R-SLAM~\cite{murai2025mast3rslam} integrates the MASt3R model~\cite{leroy2024mast3r,wang2024dust3r,duisterhof2025mastrsfm} to generate dense 3D pointmaps, which are refined using a graph-based backend. Similarly, VGGT-SLAM~\cite{maggio2025vggtslam} employs the VGGT foundation model~\cite{wang2025vggt} to align submaps, with recent extensions improving runtime through revised factor-graph formulations~\cite{maggio2026vggtslam2}. Despite these advances, end-to-end, hybrid, and foundation-model-based systems typically require substantial computational resources, limiting their deployment on mobile robotic platforms.

\subsection{Adaptive Visual Frontends}
\label{ssec:adaptive_visual_frontends}

Robust feature tracking remains challenging under variations in texture, illumination, motion blur, and sensor noise. Existing methods commonly adapt to such conditions through rules based on frontend outputs. AnyFeature-VSLAM~\cite{fontan2024anyfeature} enables different keypoint detectors and descriptors to be substituted without feature-specific manual tuning. The feature type is selected before processing; the method calibrates the selected detector against a FAST-based keypoint-count reference during initialization and adapts feature-specific matching and outlier-rejection thresholds from frontend statistics. PLPF-VSLAM~\cite{yan2023plpf} instead selects among combinations of point, line, and plane features according to the numbers of successfully matched primitives.

RL-VO~\cite{messikommer2024reinforcement} formulates the adaptation of VO decisions as a sequential decision-making problem. Its agent observes tracked keypoints, local-map statistics, and relative camera poses, and controls keyframe selection and, for SVO, the feature-grid size. However, it does not observe the raw image and therefore bases its decisions on the state produced by the frontend rather than directly on scene appearance. In contrast, our method conditions its policy on the current image and directly adapts the FAST threshold, KLT patch size, and RANSAC threshold governing feature detection, tracking, and outlier rejection.

%% file: sections/03_Methodology.tex
\section{Methodology}
\label{sec:methodology}

This section presents our image-conditioned, RL-based frontend that adapts online the feature detection and tracking parameters.

\subsection{Frontend implementation}

We build on the OpenVINS KLT frontend~\cite{geneva2020openvins}, retaining its histogram equalization, grid-based FAST detection~\cite{trajkovic1998fast}, pyramidal KLT tracking~\cite{lucas1981klt}, and fundamental-matrix RANSAC rejection. We modify the frontend to expose three parameters for per-frame control: the FAST threshold, KLT patch size, and pixel-domain RANSAC threshold $\rho$. Before RANSAC, $\rho$ is converted to normalized camera coordinates as $\rho/\max(f_x,f_y)$. All other processing and track-management settings follow OpenVINS.

\subsection{Realtime, RL-Based Parameter Tuning}
\label{ssec:rl-tuning}
The performance of the feature detection and tracking pipeline described above is highly sensitive to the choice of hyperparameters, including the FAST score threshold, KLT patch size, as well as RANSAC thresholds. These parameters jointly affect tracking robustness, accuracy, feature age, spatial coverage, and computational cost. In practice, they are typically selected by expert tuning under the assumption of a roughly known operating environment.

Recent work~\cite{messikommer2024reinforcement} has shown that RL is a suitable framework for parameter adaptation in VO pipelines. However, prior approaches rely exclusively on internal VO statistics, which fundamentally limits their ability to reason about the underlying causes of performance degradation. In contrast, the visual appearance of the scene, such as texture richness and motion blur, is the primary determinant of suitable tracking parameters.

We therefore formulate parameter tuning as a sequential decision-making problem conditioned directly on image observations, as shown in \cref{fig:figure1}. 
In particular, the observation space is composed of an encoded frame, the feature counts from the final updates of the two preceding decision steps, the most recently applied parameters, and two binary indicators denoting a new sequence and the first interaction after a reset. The image encoder is pretrained offline, as described in \cref{ssec:cnn_encoder_training}.

The action space includes three parameters: the FAST score threshold, the KLT patch size, and the RANSAC rejection threshold. During training, each action is represented by a continuous variable bounded in [-1,1], which is then mapped to the corresponding discrete frontend parameter value, as summarized in \cref{tab:parameter_space}.

A significant challenge when learning an RL policy in this context is the reward signal's strong dependency on the underlying scene. 
Empirically, we observe that the variance in rewards between easy and difficult frames severely overshadows the variance between optimal and suboptimal tracking parameters, which hinders effective learning. 
To enable the critic network to contextualize the relative difficulty of different frames, we employ a privileged critic architecture. 
Specifically, the critic is provided with the current dataset frame index, encoded as Fourier features \cite{tancik2020fourier} using 17 frequency bands.

\begin{table}[t!]
    \centering
    \setlength{\tabcolsep}{6pt}
    \caption{Action Space for the RL-Agent}
    \begin{tabularx}{\linewidth}{C|Cc}
        \toprule
        Parameter & \text{Values} & Unit \\ 
        \midrule
        FAST Threshold & $\{0, 1, \hdots, 209\}$ & \textendash \\
        KLT Patch Size & $\{3, 5, \hdots, 41\}$  & pixel \\
        RANSAC Threshold & $\{3(i/99)^2\}_{i=0}^{99}$ & pixel \\
        \bottomrule
    \end{tabularx}
    \label{tab:parameter_space}
\end{table}

\subsection{RL-Training Rewards}
\label{ssec:rl-training_rewards}
A robust sparse direct VO frontend must maintain a sufficient number of tracked features with uniform frame coverage, while minimizing the endpoint error and satisfying the available computational budget.
For this reason, we formulate the reward at step $t$ as a sum of three factors:
\begin{equation}
\begin{split}
  r^t &=  r_\text{track}^t + r_\text{cov}^t + r_\text{comp}^t
\label{eq:reward}
\end{split}
\end{equation}
where $r_\text{track}^t$, $r_\text{cov}^t$, and $r_\text{comp}^t$ denote the tracking performance, frame coverage, and computational budget rewards, respectively.

Let $\mathcal{M}^t$ denote the set of correspondences with valid, finite ground-truth optical flow. Ground-truth flow is bilinearly sampled at the feature location $\mathbf{u}_i^{t-1}$ in the preceding frame; samples outside the image domain or adjacent to invalid flow values are discarded. For a KLT estimate $\hat{\mathbf{u}}_i^t$, the endpoint error is
\begin{equation}
e_i^t = \left\|\hat{\mathbf{u}}_i^t-
\left(\mathbf{u}_i^{t-1}+\mathbf{f}^t(\mathbf{u}_i^{t-1})\right)\right\|_2.
\end{equation}
where $\mathbf{f}^t(\mathbf{u}_i^{t-1})$ is the optical flow at step $t$ evaluated at location $\mathbf{u}_i^{t-1}$.
For each $i\in\mathcal{M}^t$, we define a soft tracking quality metric
\begin{equation}
q_i^t=\exp\left(-\frac{\min(e_i^t,20)}{3}\right).
\end{equation}
The average tracking quality $Q^t$ and quality-weighted feature count $N_{\mathrm{eff}}^t$ are defined as
\begin{align}
Q^t &=
\begin{cases}
\displaystyle\frac{1}{|\mathcal{M}^t|}\sum_{i\in\mathcal{M}^t}q_i^t,
& |\mathcal{M}^t|>0,\\
0, & |\mathcal{M}^t|=0,
\end{cases}
\label{eq:tracking_quality}\\
N_{\mathrm{eff}}^t &= \sum_{i\in\mathcal{M}^t}q_i^t.
\label{eq:tracking_reward_terms}
\end{align}
From these metrics we estimate $S^t$, which penalizes shortages of reliable tracks, and $F^t$, that strongly penalizes tracking failures:
\begin{align}
S^t &= \max\left(0,\frac{30-N_{\mathrm{eff}}^t}{30}\right)
&
F^t &= \mathbf{1}\!\left[N_{\mathrm{eff}}^t<8\right]
\label{eq:tracking_reward_terms_2}
\end{align}
These terms prevent the policy from maximizing average tracking quality by retaining only a small number of highly accurate features.

The total tracking reward is expressed as:
\begin{equation}
r_\text{track}^t = {Q^t - 3(S^t)^2 - 5F^t}
\end{equation}

To encourage spatially distributed tracks, we partition the image into an $8\times8$ grid and consider only valid correspondences with $e_i^t<5$ pixels. Defining
\begin{equation}
C^t=\frac{\text{occupied cells}}{64},
\end{equation}
the coverage reward increases linearly up to $C^t=0.3$ and saturates thereafter, and it is defined as:
\begin{equation}
  r_\text{cov}^t = \min\left(\frac{C^t}{0.30},1\right)
\end{equation}
\begin{table}[t!]
    \centering
    \caption{Runtimes on AMD Ryzen 9 9950X3D CPU}
        
    \begin{tabular}{@{} l l l @{}}
    \toprule
    \textbf{Param.} & \textbf{} & \textbf{Coefficients} \\
    \midrule
    $\tau_c$ & = & $187.9201\,\mu\mathrm{s}$ \\
    \addlinespace
    $\tau_\text{klt}$ 
        & = $\nu_1 n_\text{klt}$ & $\nu_1=0.0731\,\mu\mathrm{s}/\mathrm{iter}$ \\
        & $+\, \nu_2 w^2$                     & $\nu_2=0.0166\,\mu\mathrm{s}/\mathrm{px}^2$ \\
        & $+\, \nu_3 n_\text{klt} w^2$        & $\nu_3=0.0010\,\mu\mathrm{s}/(\mathrm{iter}\,\mathrm{px}^2)$ \\
    \addlinespace
    $\tau_\text{ransac}$ 
        & = $\nu_4 n_\text{ransac}$             & $\nu_4=2.4456\,\mu\mathrm{s}/\mathrm{iter}$ \\
        & $+\, \nu_5 N_\text{ransac}$                       & $\nu_5=0.1042\,\mu\mathrm{s}/\mathrm{corr}$ \\
        & $+\, \nu_6 n_\text{ransac} N_\text{ransac}$       & $\nu_6=0.0050\,\mu\mathrm{s}/(\mathrm{iter}\,\mathrm{corr})$ \\
    \bottomrule
    \end{tabular}
    \label{tab:runtimes}
\end{table}

We additionally penalize configurations that violate the desired computational budget. 
To avoid requiring training and deployment on the same hardware, we model the frontend runtime as a function of the KLT patch size $w$, the total number of KLT iterations accumulated across all tracked features $n_\text{klt}$, the number of executed RANSAC iterations $n_\text{ransac}$, and the number of correspondences provided to RANSAC $N_\text{ransac}$. This decouples policy training from the deployment hardware, allowing the runtime model to be calibrated on one platform and the policy to be trained on another.
We estimate the tracker runtime as:
\begin{equation}
\hat{\tau}^t=\sum_{j=1}^{K_t}\left(\tau_\text{klt}^{t,j}+\tau_\text{ransac}^{t,j}+\tau_c\right),
\end{equation}
where $\tau_c$ is a constant overhead term, while $\tau_\text{klt}^{t,j}$ and $\tau_\text{ransac}^{t,j}$ denote the estimated KLT tracking and RANSAC runtimes, respectively.
The components of this model are summarized in \cref{tab:runtimes}. 

To extrapolate the reference-platform estimate to the target hardware, we use a base slowdown factor $\beta_0$. Comparing single-thread performance benchmarks \cite{cpubenchmark2026main} between the AMD Ryzen 9 9950X3D reference processor and the ARM Cortex-A57 CPU of the NVIDIA Jetson TX2 gives $\beta_0=10$. We additionally introduce a constraint multiplier $s$ and define
\begin{equation}
D_s^t=\max\left(0,
\frac{s\beta_0\hat{\tau}^t-B_\mathrm{target}}
{B_\mathrm{target}}\right),
\qquad B_\mathrm{target}=\qty{100}{ms}.
\label{eq:runtime_overrun}
\end{equation}
We train policies with $s=1$ and $s=3$. Because $\beta_0=10$, these settings are equivalent to effective reference-platform deadlines of $\qty{10}{ms}$ and $\frac{10}{3}\,\mathrm{ms}$, respectively. The $s=1$ setting represents the nominal Jetson TX2 constraint, whereas $s=3$ evaluates a three-times tighter computational setting.
The computational term is zero within the runtime budget, increases quadratically once the budget is exceeded, and is bounded by $-2$.
The computation reward is expressed as:
\begin{equation}
  r_\text{comp}^t = -0.5\min\left((D_s^t)^2,4\right)
\end{equation}
The runtime model used by the reward estimates only the frontend runtime and does not include policy inference. However, we evaluate the inference latency of our policy network on an embedded NVIDIA device, recording an average execution time of $\qty{100}{\mu s}$.

Also, we note that this training formulation requires pixel-accurate ground truth feature correspondences in order to compute drift-based rewards. Consequently, training is performed entirely in simulation. Details on the dataset and experimental setup are provided in Sec.~\ref{sec:experimental_setup}.

\subsection{CNN Encoder Training}
\label{ssec:cnn_encoder_training}

End-to-end RL directly from large images is computationally expensive, as it requires backpropagation through the visual encoder during policy optimization and storage of high-dimensional observations in the PPO rollout buffer.
To minimize runtime memory and compute, we pretrain and freeze our visual encoder using a surrogate \emph{continuous contextual bandits} task \cite{majzoubi2020efficient}. 
For encoder pretraining, we restrict the action to the KLT patch size and optimize each frame pair independently, yielding a single-step contextual-bandit problem.
Given the initial $640\times640$ grayscale frame and the sampled FAST score threshold, the policy is trained to select a KLT patch size that minimizes the average feature drift.
We optimize this surrogate task using Soft Actor-Critic (SAC).
The encoder consists of three convolution-batch-normalization-ReLU blocks with $[16,32,64]$ output channels and kernel sizes $[7,5,3]$, using a stride of $4$ in each layer. The resulting 128-dimensional vector is concatenated with the normalized FAST threshold and processed by a small MLP. The encoder input is scaled to $[0,1]$ and is not histogram-equalized.
For this single-step setup, the discount factor $\gamma$ is 0, and the reward is defined as the negative average feature drift.

%% file: sections/04_ExperimentalSetup.tex
\section{Experimental Setup}
\label{sec:experimental_setup}

This section describes the datasets used for training and evaluation, the training procedure, and the baselines against which our approach is compared.

\subsection{Datasets}
\label{ssec:datasets}

\subsubsection*{Synthetic Data}

As discussed in Sec.~\ref{ssec:rl-training_rewards}, our training reward relies on pixel-accurate ground-truth feature correspondences in order to compute feature drift. This requirement restricts policy training and feature-drift evaluation to synthetic datasets. We utilize the TartanAirV2 dataset, which provides a diverse collection of photorealistic indoor and outdoor environments, each containing multiple camera trajectories. In addition to RGB images, TartanAirV2 provides ground-truth depth maps, camera poses, and dense optical flow, providing the dense optical flow required to compute the tracking reward.

Due to the computational cost of executing the full feature detection and tracking pipeline during training, we use only a single sequence per environment for policy training. This leads to a total of 68000 frames. We evaluate on 5 sequences that are disjoint from the training set, comprising 5500 frames in total. This experiment tests the ability of the learned policy to generalize to unseen environments.

\subsubsection*{Renderer}
\label{sssec:renderer}

TartanAirV2 provides image data at a relatively low frame rate of \qty{10}{Hz}~\cite{patel2025tartanground}, which is not representative of modern machine-vision cameras commonly used in VO and VIO systems. To obtain more realistic temporal sampling, we upsample the dataset by re-rendering intermediate frames.

Given two consecutive frames $I^t$ and $I^{t+1}$, their camera poses, and depth maps, we interpolate translations using cubic splines and rotations using spherical linear interpolation. Both endpoint images are depth-warped to each interpolated pose using a pinhole model and bilinear sampling, and their valid regions are blended to form the intermediate frame.

\subsubsection*{Appearance-Regime Switching}
To robustify the model against time-varying visual conditions, we organize each trajectory as a sequence of piecewise-constant appearance regimes. We consider three regimes: nominal images,
images with motion blur, and images with both motion blur and gamma-domain sensor noise. The three image streams are generated in advance and are aligned at the frame level. A regime
switch therefore changes only the observed image. The frame index, camera motion, ground-truth optical flow, and optical-flow statistics remain unchanged.

Each parallel environment maintains an independent regime schedule. At the start of a trajectory, the loader samples an initial regime according to the distribution
\begin{equation}
  \boldsymbol{p}_{r}
  =
  \begin{bmatrix}
      p_{\mathrm{nom}} &
      p_{\mathrm{blur}} &
      p_{\mathrm{noise}}
  \end{bmatrix}.
\end{equation}
In our experiments, we use
$\boldsymbol{p}_{r}=[0.4,\,0.3,\,0.3]$.
The selected regime remains active for a number of steps sampled uniformly from
\begin{equation}
  L \sim \mathcal{U}_{\mathbb{Z}}
  \left(L_{\min},L_{\max}\right),
\end{equation}
where $L_{\min}=16$ and $L_{\max}=64$ loader steps. After $L$ steps, the loader samples a new regime. The new regime must differ from the current one.

A change in appearance does not indicate the start of a new trajectory and does not reset the visual frontend. Feature tracks and estimator state therefore persist across each transition.
This design separates appearance changes from changes in scene geometry and camera motion. It also tests whether the
frontend can adapt online without prior knowledge of the transition time or the new image condition.
  
\subsubsection*{Motion Blur}
The synthetic dataset does not contain motion blur, which is not representative of real-world camera imagery. To account for this, we augment the rendered frames with motion blur using the provided ground-truth optical flow. Specifically, we compute a per-frame motion-blur kernel from the flow field and convolve the image with this kernel to approximate the effect of continuous camera motion during exposure. We fix the simulated exposure time to $12.5\,\mathrm{ms}$, half of the $25\,\mathrm{ms}$ frame interval at the highest evaluated frame rate of $40\,\mathrm{Hz}$, which is equivalent to a shutter angle of $180^\circ$ in cinematography. We note that this model is an approximation, as it does not account for longer exposures in darker scenes.

\subsubsection*{Sensor Noise}
\begin{figure}[ht!]
    \centering
    \subfloat[Nominal]{%
        \includegraphics[width=0.3\linewidth]{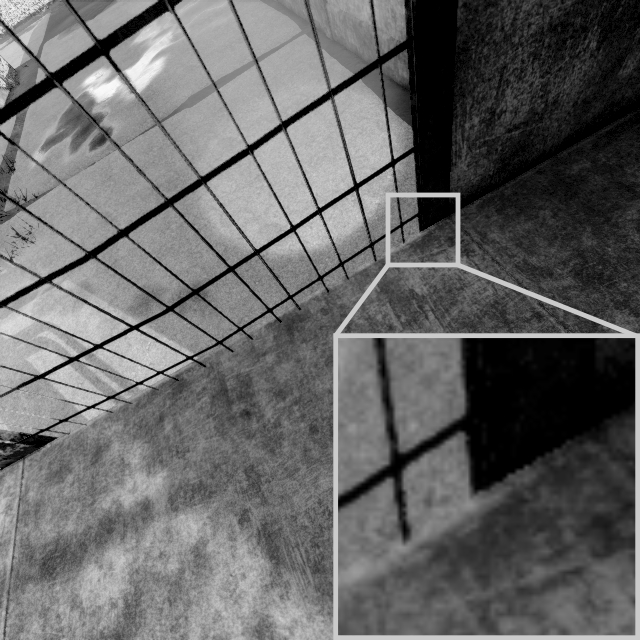}%
    }\hfill
    \subfloat[With motion-blur]{%
        \includegraphics[width=0.3\linewidth]{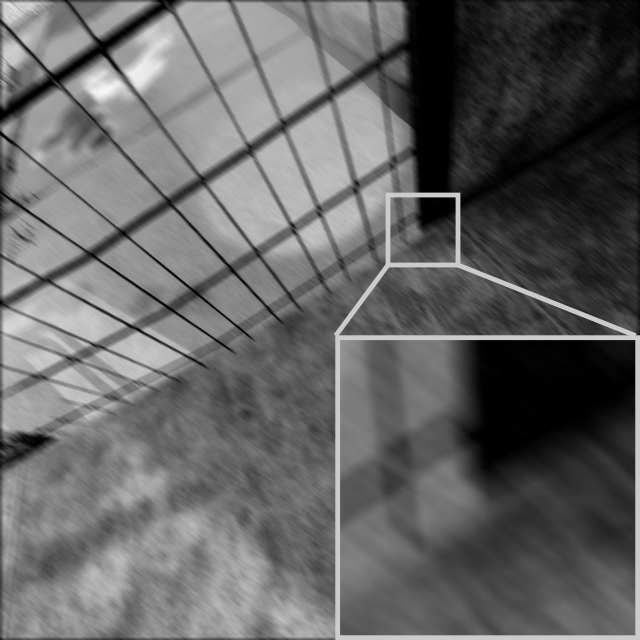}%
    }\hfill
    \subfloat[With motion-blur and noise]{%
        \includegraphics[width=0.3\linewidth]{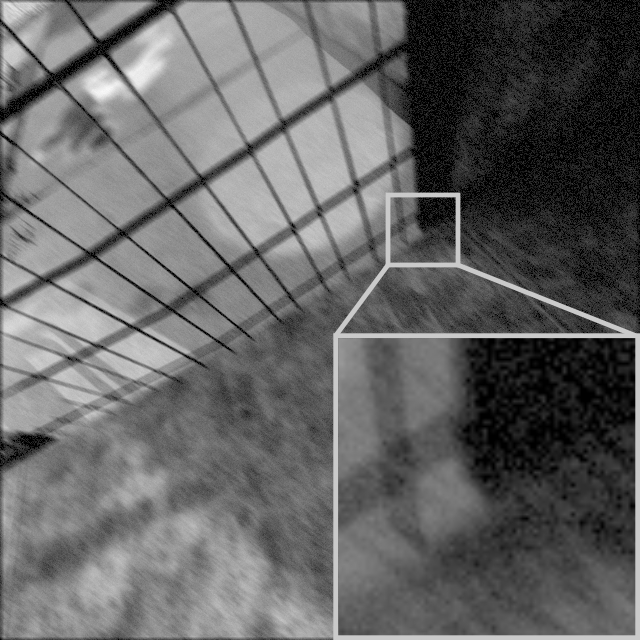}%
    }
    \caption{Since TartanAirV2 does not feature realistic noise and blur dynamics, we augment the dataset.}
    \label{fig:dataset_augmentation}
\end{figure}

Unlike real-world cameras, the dataset images are noise-free. We therefore inject synthetic sensor noise. In this work, we adopt a simple noise model that assumes additive Gaussian noise on the linear image intensities. Since the dataset provides gamma-corrected PNG images rather than raw sensor measurements, we assume a standard gamma value of $\gamma = 2.2$. Given a noise-free image $I \in [0,1]$, the noisy image $\tilde{I}$ is generated as
\begin{equation}
\tilde{I} = \left[\operatorname{clip}\left(I^\gamma + \epsilon,\,0,\,1\right)\right]^{1/\gamma},
\qquad \epsilon \sim \mathcal{N}\left(0,(0.0125)^2\right).
\end{equation}
Figure~\ref{fig:dataset_augmentation} compares nominal images with images augmented by motion blur and noise.

\subsubsection*{Real-World Datasets}
In addition to the synthetic-data experiments, we evaluate the proposed RL-based frontend on three real-world benchmark datasets: EuRoC MAV~\cite{burri2016euroc}, TUM-VI~\cite{schubert2018vidataset}, and UZH-FPV~\cite{delmerico2019uzhfpv}. We report results on all EuRoC sequences, TUM-VI \texttt{room} sequences, and UZH-FPV sequences for which ground-truth trajectories are available.

\subsection{Training Details}
\label{ssec:training_details}

Policy training is performed using proximal policy optimization (PPO). The learning rate is linearly decayed from $1 \times 10^{-4}$ to $1\times 10^{-5}$ over the course of training. Each PPO rollout contains 128 consecutive steps per environment. Training is conducted in parallel across $N = 256$ environments, yielding $256\times128=32768$ transitions per policy update. Appearance-regime switches neither terminate an episode nor reset the frontend.
The CNN encoder described in \cref{ssec:cnn_encoder_training} is frozen during RL training, while PPO optimizes the actor and critic networks. We train for $5\times10^7$ environment transitions. The PPO hyperparameters used in our experiments are summarized in \cref{tab:ppo_hyperparameters}.

\begin{table}[!t]
\centering
\caption{\textnormal{PPO Hyperparameters and Network Architecture.}}
\vspace*{-6pt}
\label{tab:ppo_hyperparameters}
\setlength{\tabcolsep}{6pt}
\begin{tabularx}{1.0\linewidth}{lX|lX}
\toprule 
\textbf{Parameter} & \textbf{Value} & \textbf{Parameter} & \textbf{Value} \\ \midrule
    discount factor $\gamma$ & $0.99$ & parallel environments & $256$ \\
    GAE-$\lambda$ & $0.95$ & rollout steps & $128$ \\
    PPO epochs & $5$ & batch size & $2048$  \\ 
    clip range & $0.2$ & policy network MLP & $[256, 256]$ \\
    entropy coefficient & $0.001$ &  value network MLP & $[256, 256]$  \\ 
\bottomrule
\end{tabularx}
\vspace*{-6pt}
\end{table}

\subsection{Baselines}
\label{ssec:baselines}

We compare our method against static parameter-selection strategies.  Unlike
the proposed image-conditioned policy, these baselines select one parameter
configuration before deployment and keep it fixed throughout each sequence.

We use particle swarm optimization (PSO)~\cite{kennedy1995particleswarmoptimization}
to optimize the same reward as the RL policy over a single static parameter
set.  \emph{PSO} is optimized on the training
split and subsequently evaluated without adaptation on the unseen
sequences.  It obeys the same train--test separation as RL and is therefore
our primary baseline. 
For the real-world datasets, we additionally compare against the parameter
configurations provided with OpenVINS~\cite{geneva2020openvins}, tuned for
the corresponding dataset or sequence.

%% file: sections/05_Results.tex
\section{Results}
\label{sec:results}

\subsection{Results on Synthetic Data}
\label{ssec:synthetic_results}

We evaluate on five TartanAirV2 sequences, not seen at training time, using the same frames and appearance-regime schedule for every method.  We consider the original \qty{10}{Hz} data and re-rendered variants at \qty{20}{Hz} and
\qty{40}{Hz}.  \Cref{tab:tartanair_complete} reports both computation
constraints: $s=1$ corresponds to a $10\,\mathrm{ms}$ reference-platform
deadline, whereas $s=3$ imposes a tighter $3.33\,\mathrm{ms}$ deadline.
\begin{table*}[t]
    \centering
    \caption{\textnormal{Complete TartanAirV2 test results.  Slash-separated entries are
    mean/median.  $N_{\mathrm{fts}}<8$ and $N_{\mathrm{fts}}<20$ count frames
    with fewer than the indicated number of raw frontend features; runtime budget violations are reported as number of frames.  Endpoint error and feature age exclude initialization frames without valid correspondences.  Bold compares RL with PSO at the same rate and constraint.}}
    \label{tab:tartanair_complete}
    \setlength{\tabcolsep}{2.6pt}
    \resizebox{\textwidth}{!}{%
    \begin{tabular}{cclrrrrrrr}
        \toprule
        $\boldsymbol{s}$ & \textbf{Rate} & \textbf{Method} &
        \makecell{\textbf{Runtime} $\boldsymbol{\downarrow}$\\$[\mathrm{ms}]$} &
        \makecell{\textbf{Endpoint error} $\boldsymbol{\downarrow}$\\$[\mathrm{px}]$} &
        \textbf{Coverage} $\boldsymbol{\uparrow}$ &
        \makecell{\textbf{Feature age} $\boldsymbol{\uparrow}$\\$[\mathrm{frames}]$} &
        \makecell{$\boldsymbol{N}_{\mathrm{fts}}<\mathbf{8}$\\$\boldsymbol{\downarrow}$} &
        \makecell{$\boldsymbol{N}_{\mathrm{fts}}<\mathbf{20}$\\$\boldsymbol{\downarrow}$} &
        \makecell{\textbf{Over budget}\\\textbf{frames} $\boldsymbol{\downarrow}$} \\
        \midrule
        \multirow{6}{*}{$1$}
        & \multirow{2}{*}{\qty{10}{Hz}} & PSO & \textbf{3.950} / 2.477 & \textbf{3.702} / 0.765 & \textbf{0.531} / \textbf{0.609} & \textbf{12.923} / \textbf{11.250} & 401 & 621 & 897 \\
        & & RL (ours) & 4.048 / \textbf{1.934} & 3.801 / \textbf{0.636} & 0.494 / 0.563 & 10.782 / 9.909 & \textbf{286} & \textbf{515} & \textbf{865} \\
        \cmidrule(lr){2-10}
        & \multirow{2}{*}{\qty{20}{Hz}} & PSO & 4.072 / 2.692 & 1.037 / 0.625 & \textbf{0.714} / \textbf{0.750} & \textbf{12.881} / \textbf{11.350} & \textbf{5} & \textbf{13} & 249 \\
        & & RL (ours) & \textbf{2.969} / \textbf{1.599} & \textbf{0.863} / \textbf{0.471} & 0.666 / 0.688 & 10.491 / 9.148 & 9 & 16 & \textbf{220} \\
        \cmidrule(lr){2-10}
        & \multirow{2}{*}{\qty{40}{Hz}} & PSO & \textbf{3.004} / \textbf{2.319} & \textbf{0.843} / 0.521 & \textbf{0.775} / \textbf{0.844} & \textbf{10.734} / 8.677 & 22 & 102 & \textbf{20} \\
        & & RL (ours) & 4.361 / 2.870 & 0.973 / \textbf{0.486} & 0.676 / 0.703 & 10.475 / \textbf{9.067} & \textbf{5} & \textbf{6} & 413 \\
        \midrule
        \multirow{6}{*}{$3$}
        & \multirow{2}{*}{\qty{10}{Hz}} & PSO & 4.836 / 2.214 & 4.554 / 0.880 & \textbf{0.637} / \textbf{0.734} & 11.253 / 9.101 & \textbf{0} & \textbf{123} & \textbf{1492} \\
        & & RL (ours) & \textbf{3.721} / \textbf{1.648} & \textbf{3.708} / \textbf{0.697} & 0.505 / 0.578 & \textbf{11.403} / \textbf{10.179} & 278 & 530 & 1553 \\
        \cmidrule(lr){2-10}
        & \multirow{2}{*}{\qty{20}{Hz}} & PSO & 2.707 / 2.053 & \textbf{1.022} / 0.553 & \textbf{0.800} / \textbf{0.875} & 10.806 / 8.731 & 17 & 65 & 1399 \\
        & & RL (ours) & \textbf{2.265} / \textbf{1.331} & 1.027 / \textbf{0.516} & 0.656 / 0.688 & \textbf{11.073} / \textbf{9.667} & \textbf{10} & \textbf{17} & \textbf{947} \\
        \cmidrule(lr){2-10}
        & \multirow{2}{*}{\qty{40}{Hz}} & PSO & 7.543 / 5.885 & \textbf{0.881} / 0.601 & \textbf{0.862} / \textbf{0.922} & 10.003 / 8.144 & \textbf{0} & \textbf{0} & 5412 \\
        & & RL (ours) & \textbf{3.023} / \textbf{2.011} & 1.168 / \textbf{0.564} & 0.639 / 0.672 & \textbf{10.513} / \textbf{8.986} & 5 & 9 & \textbf{1284} \\
        \bottomrule
    \end{tabular}%
    }
\end{table*}
Under the nominal constraint $s=1$, the clearest joint improvement occurs at
\qty{20}{Hz}.  Relative to PSO, RL reduces mean runtime by $27.1\%$,
mean EPE by $16.8\%$, and deadline violations by $11.6\%$.  Coverage and track
age are lower, indicating that the policy achieves this operating point using
fewer spatial and temporal feature redundancies.

The tight constraint $s=3$ makes the advantage at moderate and high frame
rates more pronounced.  At \qty{20}{Hz}, RL reduces runtime by $16.3\%$,
frames with fewer than 20 features ($N_{\mathrm{fts}}<20$) by $73.8\%$, and deadline violations by
$32.3\%$; mean EPE changes by only $0.5\%$, while median EPE improves by
$6.7\%$.  At \qty{40}{Hz}, RL reduces runtime by $59.9\%$ and deadline
violations by $76.3\%$.  This aggressive operating point improves median EPE
but increases mean EPE by $32.6\%$ and reduces coverage, revealing a trade-off
between typical tracking accuracy and tail behavior.

The complete results also expose two limitations.  For $s=1$ at
\qty{40}{Hz}, RL reduces low-feature frames (only six frames have fewer than
20 features, compared with 102 for PSO) but runs $45.2\%$ slower and
misses the deadline more often.  At \qty{10}{Hz} under $s=3$, RL reduces
runtime and EPE but produces lower coverage and substantially more low-feature
frames than PSO.  Thus, adaptation is most consistently beneficial at
\qty{20}{Hz}; large inter-frame motion remains challenging under a tight
computation constraint.

\subsection{Results on Real-world Sequences}
\label{ssec:real_world_results}

We next evaluate whether the frontend-level behavior transfers to downstream
state estimation.  We integrate each parameter-selection strategy into the
same monocular-inertial OpenVINS pipeline and use its dynamic inertial initializer.
Every sequence is evaluated ten times.  For RL and PSO, an action is shared across four
frontend updates and the computation-constraint multiplier is $s=1$; the
OpenVINS reference uses the dataset-specific default YAML configuration.
\begin{table*}
    \centering
    \caption{\textnormal{Monocular--inertial OpenVINS results on real-world datasets.
    Translation ATE is computed after SE(3) alignment.  Dataset-level ATE
    entries first aggregate the ten runs of each sequence and then average all
    sequences with equal weight; slash-separated entries denote the
    across-run mean/median of the indicated per-run statistic..
    Low-feature counts are pooled over all input frames and all ten runs.
    Bold identifies improvements of RL over the PSO baseline.}}
    \label{tab:real_world_openvins}
    \setlength{\tabcolsep}{4.0pt}
    \begin{tabular}{llrrrr}
        \toprule
        \textbf{Dataset} & \textbf{Method} &
        \makecell{\textbf{Mean ATE} $\boldsymbol{\downarrow}$\\$[\mathrm{m}]$} &
        \makecell{\textbf{Runtime} $\boldsymbol{\downarrow}$\\$[\mathrm{ms}]$} &
        $\boldsymbol{N}_{\mathrm{fts}}<\mathbf{8}$ &
        $\boldsymbol{N}_{\mathrm{fts}}<\mathbf{20}$ $\boldsymbol{\downarrow}$ \\
        \midrule
        \multirow{3}{*}{EuRoC}
        & PSO & 0.1531 / 0.1552 & 21.591 / 17.969 & 0 & 0 \\
        & RL (ours) & \textbf{0.1473 / 0.1460} & \textbf{9.248 / 8.369} & 0 & 10 \\
        & \cellcolor{gray!15}OpenVINS default & \cellcolor{gray!15}0.1392 / 0.1409 & \cellcolor{gray!15}9.239 / 9.013 & \cellcolor{gray!15}0 & \cellcolor{gray!15}10 \\
        \midrule
        \multirow{3}{*}{TUM-VI}
        & PSO & 0.0885 / 0.0883 & 20.114 / 16.749 & 0 & 0 \\
        & RL (ours) & \textbf{0.0839 / 0.0848} & \textbf{9.330 / 8.391} & 0 & 0 \\
        & \cellcolor{gray!15}OpenVINS default & \cellcolor{gray!15}0.0751 / 0.0758 & \cellcolor{gray!15}8.373 / 8.256 & \cellcolor{gray!15}0 & \cellcolor{gray!15}0 \\
        \midrule
        \multirow{3}{*}{UZH-FPV}
        & PSO & 0.3547 / 0.3547 & 19.409 / 15.389 & 10 & 511 \\
        & RL (ours) & \textbf{0.3237 / 0.3237} & \textbf{9.418 / 7.666} & \textbf{0} & \textbf{400} \\
        & \cellcolor{gray!15}OpenVINS default & \cellcolor{gray!15}0.3148 / 0.3148 & \cellcolor{gray!15}7.741 / 6.601 & \cellcolor{gray!15}280 & \cellcolor{gray!15}4990 \\
        \bottomrule
    \end{tabular}
\end{table*}
Relative to PSO, RL lowers the mean of the per-run mean ATE by $3.8\%$,
$5.2\%$, and $8.7\%$ on EuRoC, TUM-VI, and UZH-FPV, respectively.  The same
comparison using the mean per-run median ATE gives reductions of $7.3\%$,
$6.9\%$, and $9.5\%$.  Simultaneously, RL reduces the mean runtime by $57.2\%$, $53.6\%$, and $51.5\%$.  Thus, compared with a static
configuration optimized using the same training objective, online parameter
selection improves trajectory accuracy while requiring substantially less
work in the classical frontend.

The dataset-specific OpenVINS defaults provide a strong reference and obtain
the lowest aggregate ATE on all three datasets.  Relative to these defaults,
RL increases mean ATE by $5.8\%$, $11.7\%$, and $2.8\%$ on EuRoC, TUM-VI, and
UZH-FPV, respectively.  The UZH-FPV feature-availability results nevertheless
show a complementary robustness benefit.  RL produces no frames with fewer
than eight features, compared with 10 for PSO and 280 for the OpenVINS
default.  It produces 400 frames with fewer than 20 features, reducing these
events by $21.7\%$ relative to PSO and by $92.0\%$ relative to the
OpenVINS default.  On EuRoC and TUM-VI, low-feature events are absent or too
rare to distinguish the methods; their relevance is most apparent on the UZH-FPV sequences.

%% file: sections/06_Discussion.tex
\section{Discussion and Conclusion}
\label{sec:discussion}
\label{sec:conclusion}

The central contribution of this work is the formulation and validation of
image-conditioned RL for online frontend configuration.  At \qty{20}{Hz} under
the tight constraint, RL reduces runtime by $16.3\%$, frames with fewer than
20 features by $73.8\%$, and deadline violations by $32.3\%$ relative to
PSO, with only a $0.5\%$ change in mean EPE. 

Zero-shot transfer provides the strongest evidence for the policy's utility.
Although trained only on TartanAirV2, RL reduces mean ATE over PSO by
$3.8$--$8.7\%$ across EuRoC, TUM-VI, and UZH-FPV, while reducing native
VIO-step time by $51.5$--$57.2\%$.  

%% file: sections/A_Acknowledgements.tex
\section*{Acknowledgments}
This work was supported by the European Union’s Horizon Europe Research and Innovation Programme under grant agreement No. 101120732 (AUTOASSESS) and the European Research Council (ERC) under grant agreement No. 864042 (AGILEFLIGHT).
This research was carried out at the Jet Propulsion Laboratory, California Institute of Technology, and was sponsored by JVSRP and the National Aeronautics and Space Administration (80NM0018D0004). ©2026. All rights reserved.

%% file: root.bbl
\begin{thebibliography}{10}
\providecommand{\url}[1]{#1}
\csname url@samestyle\endcsname
\providecommand{\newblock}{\relax}
\providecommand{\bibinfo}[2]{#2}
\providecommand{\BIBentrySTDinterwordspacing}{\spaceskip=0pt\relax}
\providecommand{\BIBentryALTinterwordstretchfactor}{4}
\providecommand{\BIBentryALTinterwordspacing}{\spaceskip=\fontdimen2\font plus
\BIBentryALTinterwordstretchfactor\fontdimen3\font minus \fontdimen4\font\relax}
\providecommand{\BIBforeignlanguage}[2]{{%
\expandafter\ifx\csname l@#1\endcsname\relax
\typeout{** WARNING: IEEEtran.bst: No hyphenation pattern has been}%
\typeout{** loaded for the language `#1'. Using the pattern for}%
\typeout{** the default language instead.}%
\else
\language=\csname l@#1\endcsname
\fi
#2}}
\providecommand{\BIBdecl}{\relax}
\BIBdecl

\bibitem{carlone2026slam}
L.~Carlone, A.~Kim, T.~D. Barfoot, D.~Cremers, and F.~Dellaert, Eds., \emph{{SLAM Handbook: From Localization and Mapping to Spatial Intelligence}}.\hskip 1em plus 0.5em minus 0.4em\relax Cambridge University Press, 2026.

\bibitem{tranzatto2024team}
M.~Tranzatto, M.~Dharmadhikari, L.~Bernreiter, M.~Camurri, S.~Khattak, F.~Mascarich, P.~Pfreundschuh, D.~Wisth, S.~Zimmermann, M.~Kulkarni \emph{et~al.}, ``Team cerberus wins the darpa subterranean challenge: Technical overview and lessons learned,'' \emph{Field Robotics}, vol.~4, pp. 349--312, 2024.

\bibitem{zhao2024subt}
S.~Zhao, Y.~Gao, T.~Wu, D.~Singh, R.~Jiang, H.~Sun, M.~Sarawata, Y.~Qiu, W.~Whittaker, I.~Higgins \emph{et~al.}, ``Subt-mrs dataset: Pushing slam towards all-weather environments,'' in \emph{Proceedings of the IEEE/CVF conference on computer vision and pattern recognition}, 2024, pp. 22\,647--22\,657.

\bibitem{cheng2005visual}
Y.~Cheng, M.~Maimone, and L.~Matthies, ``Visual odometry on the mars exploration rovers,'' in \emph{2005 IEEE International Conference on Systems, Man and Cybernetics}, vol.~1.\hskip 1em plus 0.5em minus 0.4em\relax IEEE, 2005, pp. 903--910.

\bibitem{delaune2020xvio}
J.~Delaune, D.~S. Bayard, and R.~Brockers, ``xvio: A range-visual-inertial odometry framework,'' 2020.

\bibitem{mourikis2007msckf}
A.~I. Mourikis and S.~I. Roumeliotis, ``A multi-state constraint kalman filter for vision-aided inertial navigation,'' in \emph{Proceedings of the IEEE International Conference on Robotics and Automation (ICRA)}, 2007, pp. 3565--3572.

\bibitem{forster2014svo}
C.~Forster, M.~Pizzoli, and D.~Scaramuzza, ``{SVO}: Fast semi-direct monocular visual odometry,'' in \emph{ICRA}, 2014.

\bibitem{bloesch2015robust}
M.~Bloesch, S.~Omari, M.~Hutter, and R.~Siegwart, ``Robust visual inertial odometry using a direct ekf-based approach,'' in \emph{2015 IEEE/RSJ international conference on intelligent robots and systems (IROS)}.\hskip 1em plus 0.5em minus 0.4em\relax IEEE, 2015, pp. 298--304.

\bibitem{bloesch2017iterated}
M.~Bloesch, M.~Burri, S.~Omari, M.~Hutter, and R.~Siegwart, ``Iterated extended kalman filter based visual-inertial odometry using direct photometric feedback,'' \emph{The International Journal of Robotics Research}, vol.~36, no.~10, pp. 1053--1072, 2017.

\bibitem{leutenegger2015keyframe}
S.~Leutenegger, S.~Lynen, M.~Bosse, R.~Siegwart, and P.~Furgale, ``Keyframe-based visual--inertial odometry using nonlinear optimization,'' \emph{The International Journal of Robotics Research}, vol.~34, no.~3, pp. 314--334, 2015.

\bibitem{geneva2020openvins}
P.~Geneva, K.~Eckenhoff, W.~Lee, Y.~Yang, and G.~Huang, ``Openvins: A research platform for visual-inertial estimation,'' in \emph{2020 IEEE International Conference on Robotics and Automation (ICRA)}.\hskip 1em plus 0.5em minus 0.4em\relax IEEE, 2020, pp. 4666--4672.

\bibitem{lucas1981klt}
B.~D. Lucas and T.~Kanade, ``An iterative image registration technique with an application to stereo vision,'' in \emph{IJCAI}, 1981.

\bibitem{wang2020tartanair}
W.~Wang, D.~Zhu, X.~Wang \emph{et~al.}, ``Tartanair: A dataset to push the limits of visual slam,'' in \emph{IROS}, 2020.

\bibitem{patel2025tartanground}
M.~Patel, F.~Yang, Y.~Qiu, C.~Cadena, S.~Scherer, M.~Hutter, and W.~Wang, ``Tartanground: A large-scale dataset for ground robot perception and navigation,'' \emph{arXiv preprint arXiv:2505.10696}, 2025.

\bibitem{burri2016euroc}
\BIBentryALTinterwordspacing
M.~Burri, J.~Nikolic, P.~Gohl, T.~Schneider, J.~Rehder, S.~Omari, M.~W. Achtelik, and R.~Siegwart, ``The euroc micro aerial vehicle datasets,'' \emph{Int. J. Rob. Res.}, vol.~35, no.~10, p. 1157–1163, Sep. 2016. [Online]. Available: \url{https://doi.org/10.1177/0278364915620033}
\BIBentrySTDinterwordspacing

\bibitem{delmerico2019uzhfpv}
J.~Delmerico, T.~Cieslewski, H.~Rebecq, M.~Faessler, and D.~Scaramuzza, ``Are we ready for autonomous drone racing? the uzh-fpv drone racing dataset,'' in \emph{2019 International Conference on Robotics and Automation (ICRA)}.\hskip 1em plus 0.5em minus 0.4em\relax IEEE, 2019, pp. 6713--6719.

\bibitem{fontan2024anyfeature}
A.~Fontan, J.~Civera, and M.~Milford, ``{AnyFeature-VSLAM}: Automating the usage of any chosen feature into visual slam,'' in \emph{Robotics: Science and Systems}, 2024.

\bibitem{yan2023plpf}
J.~Yan, Y.~Zheng, J.~Yang, L.~Mihaylova, W.~Yuan, and F.~Gu, ``Plpf‐vslam: An indoor visual slam with adaptive fusion of point‐line‐plane features,'' \emph{Journal of Field Robotics}, 2023.

\bibitem{messikommer2024reinforcement}
N.~Messikommer, G.~Cioffi, M.~Gehrig, and D.~Scaramuzza, ``Reinforcement learning meets visual odometry,'' in \emph{ECCV}, 2024.

\bibitem{schubert2018vidataset}
D.~Schubert, T.~Goll, N.~Demmel, V.~Usenko, J.~Stueckler, and D.~Cremers, ``The tum vi benchmark for evaluating visual-inertial odometry,'' in \emph{International Conference on Intelligent Robots and Systems (IROS)}, October 2018.

\bibitem{kim2022proprietary}
P.~Kim, J.~Kim, M.~Song, Y.~Lee, M.~Jung, and H.-G. Kim, ``A benchmark comparison of four off-the-shelf proprietary visual--inertial odometry systems,'' \emph{Sensors}, vol.~22, no.~24, p. 9873, 2022.

\bibitem{cadena2016slam}
C.~Cadena, L.~Carlone, H.~Carrillo, Y.~Latif, D.~Scaramuzza, J.~Neira, I.~D. Reid, and J.~J. Leonard, ``Simultaneous localization and mapping: Present, future, and the robust-perception age,'' \emph{CoRR}, 2016.

\bibitem{zhou2015guidance}
G.~Zhou, L.~Fang, K.~Tang, H.~Zhang, K.~Wang, and K.~Yang, ``Guidance: A visual sensing platform for robotic applications,'' in \emph{CVPR Workshops}, 2015.

\bibitem{kerl2015dense}
C.~Kerl, J.~Stueckler, and D.~Cremers, ``Dense continuous-time tracking and mapping with rolling shutter rgb-d cameras,'' in \emph{ICCV}, 2015.

\bibitem{engel2018directsparse}
J.~Engel, V.~Koltun, and D.~Cremers, ``Direct sparse odometry,'' \emph{IEEE Transactions on Pattern Analysis and Machine Intelligence}, 2018.

\bibitem{murartal2015orb}
R.~Mur-Artal, J.~M.~M. Montiel, and J.~D. Tardos, ``Orb-slam: A versatile and accurate monocular slam system,'' \emph{IEEE Transactions on Robotics}, 2015.

\bibitem{murartal2017orbslam2}
R.~Mur-Artal and J.~D. Tardos, ``Orb-slam2: An open-source slam system for monocular, stereo, and rgb-d cameras,'' \emph{IEEE Transactions on Robotics}, 2017.

\bibitem{campos2021orbslam3}
C.~Campos, R.~Elvira, J.~J.~G. Rodriguez, J.~M.~M. Montiel, and J.~D. Tardos, ``Orb-slam3: An accurate open-source library for visual, visual–inertial, and multimap slam,'' \emph{IEEE Transactions on Robotics}, 2021.

\bibitem{detone2018superpoint}
D.~DeTone, T.~Malisiewicz, and A.~Rabinovich, ``Superpoint: Self-supervised interest point detection and description,'' in \emph{CVPR Workshops}, 2018.

\bibitem{lindenberger2023lightglue}
P.~Lindenberger, P.-E. Sarlin, and M.~Pollefeys, ``Lightglue: Local feature matching at light speed,'' in \emph{ICCV}, 2023.

\bibitem{sarlin2020superglue}
P.-E. Sarlin, D.~DeTone, T.~Malisiewicz, and A.~Rabinovich, ``Superglue: Learning feature matching with graph neural networks,'' in \emph{CVPR}, 2020.

\bibitem{wang2017deepvo}
S.~Wang, R.~Clark, H.~Wen, and N.~Trigoni, ``Deepvo: Towards end-to-end visual odometry with deep recurrent convolutional neural networks,'' in \emph{ICRA}, 2017.

\bibitem{wang2020tartanvo}
W.~Wang, Y.~Hu, and S.~Scherer, ``Tartanvo: A generalizable learning-based vo,'' in \emph{CoRL}, 2020.

\bibitem{teed2021droid}
Z.~Teed and J.~Deng, ``Droid-slam: Deep visual slam for monocular, stereo, and rgb-d cameras,'' \emph{NeurIPS}, 2021.

\bibitem{teed2023dpvo}
\BIBentryALTinterwordspacing
Z.~Teed, L.~Lipson, and J.~Deng, ``Deep patch visual odometry,'' in \emph{Advances in Neural Information Processing Systems}, A.~Oh, T.~Naumann, A.~Globerson, K.~Saenko, M.~Hardt, and S.~Levine, Eds., vol.~36.\hskip 1em plus 0.5em minus 0.4em\relax Curran Associates, Inc., 2023, pp. 39\,033--39\,051. [Online]. Available: \url{https://proceedings.neurips.cc/paper_files/paper/2023/file/7ac484b0f1a1719ad5be9aa8c8455fbb-Paper-Conference.pdf}
\BIBentrySTDinterwordspacing

\bibitem{lipson2024deeppatch}
L.~Lipson, Z.~Teed, and J.~Deng, ``Deep patch visual slam,'' in \emph{ECCV}, 2024.

\bibitem{murai2025mast3rslam}
R.~Murai, E.~Dexheimer, and A.~J. Davison, ``Mast3r-slam: Real-time dense slam with 3d reconstruction priors,'' in \emph{CVPR}, 2025.

\bibitem{leroy2024mast3r}
V.~Leroy, Y.~Cabon, and J.~Revaud, ``Grounding image matching in 3d with mast3r,'' in \emph{ECCV}, 2024.

\bibitem{wang2024dust3r}
S.~Wang, V.~Leroy, Y.~Cabon \emph{et~al.}, ``Dust3r: Geometric 3d vision made easy,'' in \emph{CVPR}, 2024.

\bibitem{duisterhof2025mastrsfm}
B.~P. Duisterhof, L.~Zust, P.~Weinzaepfel, V.~Leroy, Y.~Cabon, and J.~Revaud, ``{MAS}t3r-sfm: a fully-integrated solution for unconstrained structure-from-motion,'' in \emph{International Conference on 3D Vision 2025}, 2025.

\bibitem{maggio2025vggtslam}
D.~Maggio, H.~Lim, and L.~Carlone, ``Vggt-slam: Dense rgb slam optimized on the sl(4) manifold,'' \emph{NeurIPS}, 2025.

\bibitem{wang2025vggt}
J.~Wang, M.~Chen, N.~Karaev \emph{et~al.}, ``Vggt: Visual geometry grounded transformer,'' in \emph{CVPR}, 2025.

\bibitem{maggio2026vggtslam2}
D.~Maggio and L.~Carlone, ``Vggt-slam 2.0: Real-time dense feed-forward scene reconstruction,'' 2026.

\bibitem{trajkovic1998fast}
M.~Trajković and M.~Hedley, ``Fast corner detection,'' \emph{Image and Vision Computing}, vol.~16, no.~2, pp. 75--87, 1998.

\bibitem{tancik2020fourier}
M.~Tancik, P.~Srinivasan, B.~Mildenhall, S.~Fridovich-Keil, N.~Raghavan, U.~Singhal, R.~Ramamoorthi, J.~Barron, and R.~Ng, ``Fourier features let networks learn high frequency functions in low dimensional domains,'' \emph{Advances in neural information processing systems}, vol.~33, pp. 7537--7547, 2020.

\bibitem{cpubenchmark2026main}
{PassMark Software}, ``Passmark software - pc benchmark charts,'' \url{https://www.cpubenchmark.net/}, 2026, accessed: 2026-03-05.

\bibitem{majzoubi2020efficient}
M.~Majzoubi, C.~Zhang, R.~Chari, A.~Krishnamurthy, J.~Langford, and A.~Slivkins, ``Efficient contextual bandits with continuous actions,'' \emph{Advances in Neural Information Processing Systems}, vol.~33, pp. 349--360, 2020.

\bibitem{kennedy1995particleswarmoptimization}
J.~Kennedy and R.~Eberhart, ``Particle swarm optimization,'' in \emph{Proceedings of ICNN'95 - International Conference on Neural Networks}, vol.~4, 1995, pp. 1942--1948.

\end{thebibliography}
